%% file: main.tex
\documentclass[letterpaper, 10 pt, conference]{ieeeconf}  

\IEEEoverridecommandlockouts                              

\usepackage[T1]{fontenc}
\usepackage[utf8]{inputenc} 

\usepackage{balance}
\usepackage{url}
\usepackage{amsmath}
\usepackage{amssymb}
\usepackage{graphicx}
\usepackage[ruled,linesnumbered,boxed,noend]{algorithm2e}
\usepackage{booktabs}
\usepackage{graphicx}
\usepackage{threeparttable}
\usepackage{stfloats}
\usepackage{setspace}
\usepackage{tabularx}
\usepackage{float}
\usepackage{booktabs}
\usepackage{subcaption}
\usepackage[font={small}]{caption}
\usepackage{enumerate}
\usepackage{makecell}
\usepackage{threeparttable}
\usepackage{hyperref}

\title{\LARGE \bf}
\title{\LARGE \bf
Perception-aware Planning for Quadrotor Flight in Unknown and Feature-limited Environments  }

\author{Chenxin Yu$^{1,*}$, Zihong Lu$^{1,*}$, Jie Mei$^1$, Boyu Zhou$^{2,3,\dagger}$ 
\thanks{$^*$ indicates equal contribution. $^\dagger$ indicates Corresponding Author.}
\thanks{$^1$ School of Intelligence Science and Engineering, Harbin Institute of Technology, Shenzhen, China (e-mail: studyxinchen@gmail.com; luzong2001@gmail.com; jmei@hit.edu.cn). }
\thanks{$^2$ Department of Mechanical and Energy Engineering, Southern University of Science and Technology, Shenzhen, China (e-mail: zhouby@sustech.edu.cn).}
\thanks{$^3$ Differential Robotics.}
\thanks{Project supported by the Young Scientists Fund of the National Natural Science Foundation of China (Grant No. 62403502).}
}

\begin{document}

\maketitle
\thispagestyle{empty}
\pagestyle{empty}

\begin{abstract}

Various studies on perception-aware planning
have been proposed to enhance the state estimation accuracy of
quadrotors in visually degraded environments. However, many
existing methods heavily rely on prior environmental knowledge and face significant limitations in previously unknown environments with sparse localization features, which greatly limits their practical application. In this paper, we present a perception-aware planning method for quadrotor flight in unknown and feature-limited environments that properly allocates perception resources among environmental information during navigation. We introduce a viewpoint transition graph that allows for the adaptive selection of local target viewpoints, which guide the quadrotor to efficiently navigate to the goal while maintaining sufficient localizability and without being trapped in feature-limited regions. During the local planning, a novel yaw trajectory generation method that simultaneously considers exploration capability and localizability is presented. It constructs a localizable corridor via feature co-visibility evaluation to ensure localization robustness in a computationally
efficient way. Through validations conducted in both simulation and real-world experiments, we demonstrate the feasibility and real-time performance of the proposed method. The source code is released for the reference of the community\footnote{\url{https://github.com/Robotics-STAR-Lab/LA-Planner}}.


\end{abstract}

\input{sections/introduction.tex}
\input{sections/related_work.tex}
\input{sections/problem_formulation.tex}
\input{sections/system_overview.tex}

\input{sections/algorithm.tex}
\input{sections/results.tex}

\input{sections/conclusion.tex}
{
    \balance
    \bibliographystyle{IEEEtran}
    \bibliography{IEEEabrv, bib/bibliography}
}

\end{document}

%% file: sections/introduction.tex
\section{Introduction}
\label{sec:introduction}



Visual-inertial odometry (VIO) has been a popular method for state estimation of unmanned aerial vehicles (UAVs). However, even well-established methods \cite{qin2019general,campos2021orb}, struggle in feature-scarce conditions such as textureless surfaces and low light. To mitigate localization failures, researchers have carried out numerous studies of \textit{perception-aware planning}, incorporating perception constraints into trajectory generation to enhance localizability.

\begin{figure}[t]
    \centering
    \includegraphics[width=\columnwidth]{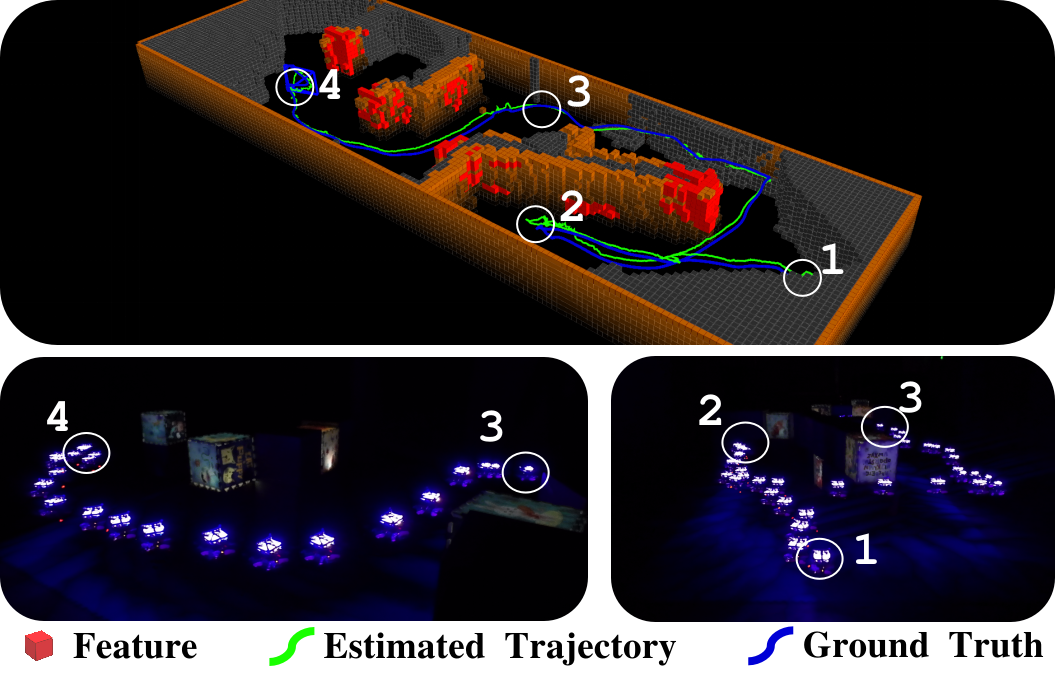} 
    \caption{Real-world flight experiment in an unknown, feature-limited, and dark environment. A dead-end is designed to test the UAV's ability to find a feasible path in a complex environment while ensuring sufficient feature visibility to maintain localization. Our method enables the UAV to successfully navigate with high efficiency and low state estimation error.}
    \label{fig:real}
    \vspace{-0.6cm}
\end{figure}


Although traditional perception-aware planners perform well with prior environmental knowledge, such as feature and obstacle locations, they face significant limitations in previously unknown environments with sparse localization features. Firstly, current methods lack a systematic mechanism to allocate perception resources among environmental information properly. Many UAVs typically have a restricted field of view (FOV). When navigating through such environments, they must simultaneously track visual features for localization and observe unknown regions to progress towards a distant goal. However, existing methods usually prioritize localizability by maximizing feature-related information. As a result, the UAV's FOV can be severely restricted by the visual features, making it difficult to observe unknown areas and navigate forward. Moreover, this over-emphasis on features may cause the UAV to overlook obstacles along its flight path, leading to potential collisions. Secondly, there is a lack of a computationally efficient way to integrate incrementally discovered features and unknown regions into trajectory planning. To assess the localization quality of a single pose, the visibility of numerous features needs to be evaluated, and this process must be repeated multiple times in trajectory planning, which incurs significant computation. This is unfavorable for navigation in unknown environments that require frequent replanning.

To address the challenges mentioned above, we present a novel perception-aware planning method for quadrotor flight in unknown and feature-limited environments in this paper. First, we introduce a viewpoint transition graph that allows for the adaptive selection of local target viewpoints. These viewpoints guide the UAV to efficiently navigate to the goal while maintaining sufficient localizability and without being trapped in feature-limited regions. Specifically, the graph captures two kinds of viewpoints: those with enough observable features for localization, and those near the unknown area's frontier, which are capable of observing both the frontier and enough features. An edge links two viewpoints when there are sufficient features to ensure localizability during the transition. By employing this, a search algorithm identifies local target viewpoints with high potential to efficiently reach the goal. At the trajectory generation level, we overcome the limitations of conventional planning methods: their complexity scales with feature counts and they tend toward over-optimization of localizability. Specifically, for each trajectory segment, we generate a localizable corridor via feature co-visibility evaluation, which then serves as a concise localizability constraint. Through optimization, we tend to enlarge the amount of information obtained from unknown areas while maintaining the necessary localization capabilities. Our approach systematically enables faster and safer navigation while maintaining sufficient localizability. Additionally, it allows for faster replanning in complex unknown environments.

Our method mitigates visual degradation while ensuring sufficient exploration capability in unknown and feature-limited environments, enabling efficient perception-aware navigation through such environments. The proposed approach is validated through both simulation and real-world experiments. The contributions of this work are as follows:
\begin{enumerate}
    \item A viewpoint transition graph that adaptively selects local targets, balancing localization and navigation efficiency.
    \item A local planner that employs a localizable corridor-based method to efficiently ensure localization robustness, and allocates perception between known and unknown areas more properly.

    \item Comprehensive validation through simulation and real-world experiments, demonstrating feasibility, stability, and real-time performance.
\end{enumerate}

%% file: sections/related_work.tex
\section{Related Work}
\label{sec:related_work}

\subsection{Perception-Aware Planning}
\label{subsec:perception_aware_planning}


In recent decades, numerous studies on perception-aware planning have been proposed for quadrotor flights in visually degraded environments. Traditional methods guide UAVs to avoid feature-limited regions by evaluating information gain. Zhang et al. \cite{zhang2018perception}-\cite{zhang2020fisher} utilized fisher information to quantify the localizability of the generated trajectories. Bartolomei et al. \cite{bartolomei2020perception} utilized semantic segmentation to identify and avoid texture-less areas on the map. Approaches that evaluate feature matchability to improve the accuracy of state estimation more effectively are also proposed. Falanga et al. \cite{falanga2018pampc} presented a perception-aware model predictive control algorithm for quadrotors, which maximizes the visibility of a point of interest and minimizes the velocity of its projection. Spasejevic et al. \cite{spasojevic2020perception} proposed an efficient time optimal path parameterization algorithm that preserves a given set of features within the FOV. Furthermore, some approaches model the feature matchability as a soft constraint in the optimizing problem. Murali et al. \cite{murali2019perception} presented an addition to the optimal control problem with predetermined trajectory that ensures both feature co-visibility and motion agility. Chen et al. \cite{chen2024apace} adopted a two-stage strategy to generate perception-aware trajectories leveraging a novelty decomposable visibility model.  

However, the aforementioned works prioritize localizability and overlook the observation of unknown areas. When deployed in previously unknown environments, these methods will over-emphasize maximizing feature-related metrics. As a result, their FOV will be severely restricted by the currently visible features. This significantly weakens their exploration ability of unknown areas and hinders efficient navigation in such environments. Additionally, some of these methods \cite{murali2019perception} \cite{chen2024apace} consume a large amount of computational resources to evaluate the visibility of features to ensure localizability. As the number of features in the environment increases, the complexity of the algorithm also rises sharply, leading to a time-consuming trajectory generation. This prevents them from performing high-frequency replanning in complex unknown environments.



\subsection{Navigation in Unknown Environments}
\label{subsec:navigation_in_unknown_environments}

Traditional navigation methods typically adopt a receding horizon approach, continuously generating local trajectories to navigate toward the goal in unknown environments. Zhou et al. \cite{zhou2019robust} proposed a robust quadrotor planning system that integrates kinodynamic search, B-spline optimization, and time adjustment to enable fast and safe flight. Another study \cite{zhou2020ego} introduced an ESDF-free gradient-based planner, which reduces computational overhead by storing obstacle information only when necessary. Oleynikova et al. \cite{oleynikova2018safe} enhanced goal selection by incorporating exploration gain and global goal guidance. However, these methods assume that reliable localization is inherently available and do not consider localization constraints, making them unsuitable for navigation in unknown and feature-limited environments.

%% file: sections/problem_formulation.tex
\section{Problem Formulation}
\label{sec:problem_formulation}

Given start and goal positions in a feature-limited environment, we aim to generate continuous trajectories and guide the quadrotor to reach the goal efficiently without any prior environmental information. The quadrotor is only allowed to use its onboard camera and inertial measurement unit (IMU) as sensors for perception and state estimation. The FOV of the camera is limited. Therefore, the quadrotor must simultaneously track features for localization and observe unknown regions to complete the navigation. 

During the navigation, the constraints of safety, dynamic feasibility, and localizability must be strictly satisfied. Specifically, the following situations will be judged as failures:

\begin{enumerate}
    \item Collides with obstacles.
    \item Fails to localize due to excessive state estimation error.
    \item Fails to complete the mission within the specified time.
\end{enumerate}

%% file: sections/system_overview.tex
\section{System Overview}
\label{sec:system_overview}

 \begin{figure}[tbp]
    \centering
    \includegraphics[width=8.5cm]{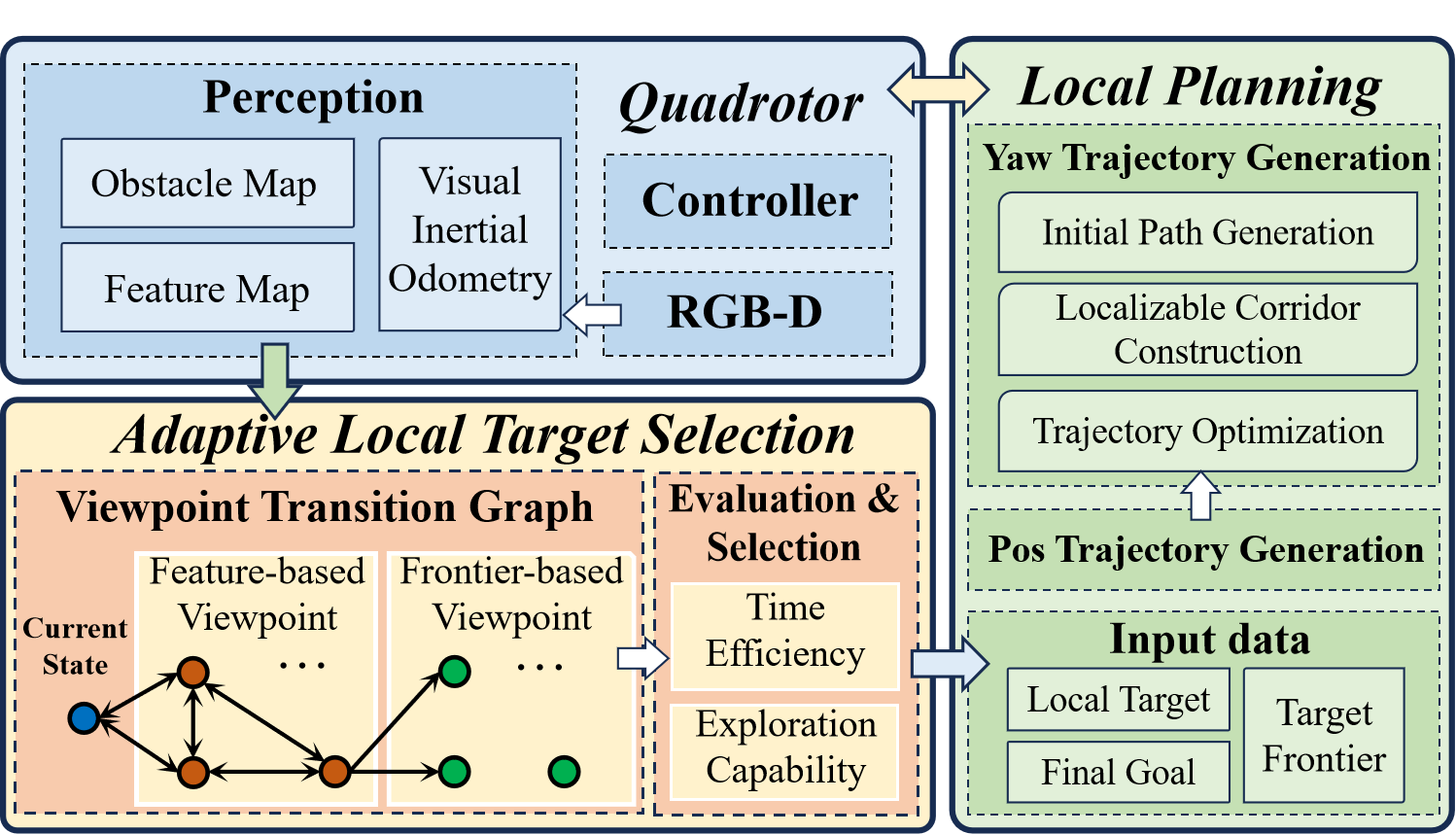}  
    \caption{An overview of the proposed perception-aware navigation framework, which consists of three main modules: 1) A perception module that provides state estimation and map information; 2) A local target selector using a viewpoint transition graph (\ref{subsec:viewpoint transition graph}) to adaptively select targets for efficient navigation while maintaining sufficient localizability. 3) A local planner that employs a localizable corridor (\ref{subsec:yaw localizable corridor generation}) to generate localizable reference trajectories toward the local targets.}
    \label{fig:system_overview}
    \vspace{-0.5cm}
\end{figure}

The system architecture is illustrated in Fig.\ref{fig:system_overview}. The quadrotor is equipped with an RGB-D camera and an IMU for perception, and its state estimation is performed by a feature-based VIO. A viewpoint transition graph (\ref{subsec:viewpoint transition graph}) is employed to adaptively select potential local targets. The graph consists of the UAV’s current state and two types of viewpoints, which are based on the boundary between known and unknown regions and features. An evaluation module (\ref{subsec:viewpoint evaluation and selection}) then assesses each local target based on time efficiency and exploration capability and selects the one with the highest score. The selected local target and its observed frontier cells are then passed to the local planner. Utilizing the differential flatness property of the quadrotor \cite{mellinger2011minimum}, the local planning problem is decomposed into position and yaw trajectory generation. In \ref{subsec:position trajectory generation}, a position trajectory is generated to guide the quadrotor toward the local target. After that, a localizable corridor-based method is employed to generate a yaw trajectory to efficiently ensure localization robustness without numerous visibility evaluations during optimization (\ref{subsec:yaw trajectory generation}). The system executes the above process in a loop until the quadrotor reaches the final goal. 




%% file: sections/algorithm.tex
\section{Methodology}
\label{sec:methodology}

\subsection{Adaptive Local Target Selection}
\label{subsec:viewpoint generation}

The local target should guide the UAV to efficiently navigate toward the goal while maintaining sufficient localizability and without being trapped in feature-limited regions. The local target selection involves a viewpoint transition graph, which captures two types of viewpoints, and includes an evaluation and selection process to determine the local target.

\subsubsection{Viewpoint Transition Graph}
\label{subsec:viewpoint transition graph}
\begin{figure}[t]
    \centering
    \includegraphics[width=\columnwidth]{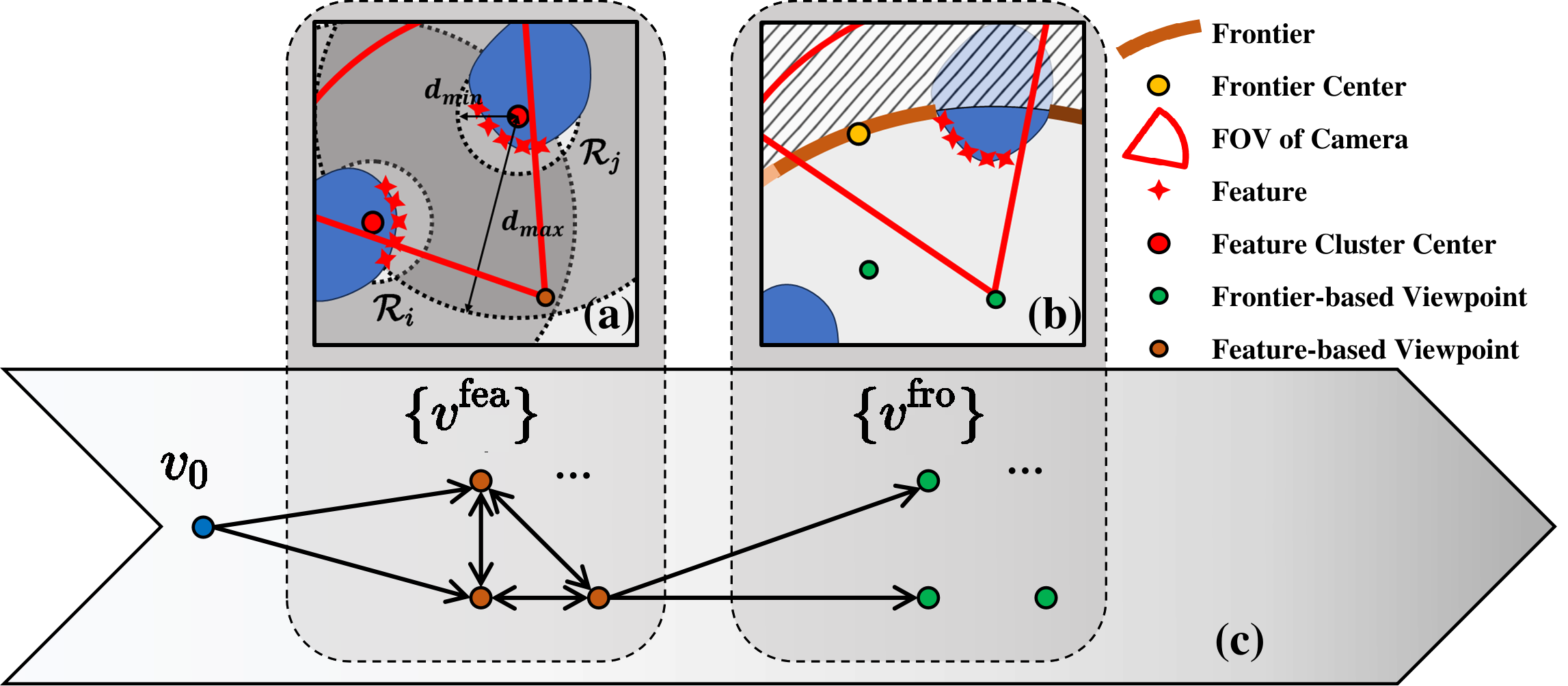} 
    \caption{Illustration of the viewpoint transition graph. (a) An example of a feature-based viewpoint selected at the intersections of regions surrounding feature cluster centers to observe both clusters. (b) An example of a frontier-based viewpoint, observing sufficient frontier cells and features. (c) Graph structure, where unidirectional arrows denote unidirectional edges, and bidirectional arrows indicate bidirectional edges.}
    \label{fig:viewpoint_graph}
    \vspace{-0.8cm}
\end{figure}

Finding a suitable local target in feature-limited regions is challenging as it must ensure localizability while also being reachable with sufficient features along the path. To address this, we introduce a viewpoint transition graph. This graph includes frontier-based viewpoints, which can observe enough unknown regions along with sufficient features, allowing them to guide the UAV step by step toward the goal. We also generate feature-based viewpoints to assess whether these frontier-based viewpoints can be reached from the current state while maintaining sufficient localizability along the path. The viewpoint transition graph is shown in Fig.\ref{fig:viewpoint_graph}.

We define each viewpoint in the graph as v = \((p, \psi)\), where \( p \) represents the position of the viewpoint, and \( \psi \) is the yaw angle that determines the observation direction. Since the quadrotor is a differentially flat system, both \( p \) and \( \psi \) are flat outputs.

\textit{\textbf{Frontier-based Viewpoints}}: To generate viewpoints capable of observing unknown regions, which represent the exploration capability of drones, we sample frontier-based viewpoints near the boundaries between known and unknown areas. These boundaries are known as frontiers. Following the method in \cite{zhou2021fuel}, we first cluster these frontiers into groups \( {FC}_1, {FC}_2, \dots, {FC}_n \). For each frontier cluster \( {FC}_i \), multiple viewpoints are sampled and the \( j \)-th viewpoint associated with \( {FC}_i \) is defined as:  
\(
v_{i,j}^{\text{fro}} = (p_{i,j}^{\text{fro}}, \psi_{i,j}^{\text{fro}})
\). The positions of viewpoints are sampled in cylindrical coordinates:
\begin{equation}
\begin{aligned}
p_{i,j}^{\text{fro}} &= \overline{{FC}_i} + r_k \left[\cos \theta_j, \sin \theta_j, 0\right]^{T} + \left[0, 0, z_m\right]^{T}
\end{aligned}
\end{equation}
where \( \overline{{FC}_i} \) is the centroid of \( {FC}_i \), \( r_k \) is the sample radius, \( \theta_j \in [-\pi, \pi] \) is the yaw sampling angle, and \( z_m \) is the height offset.
Each viewpoint \( v_{i,j}^{\text{fro}} \) should satisfy the following conditions:
\begin{equation}
\sum_{l \in {L}} \mathbb{I}(l \in \text{FOV}(v_{i,j}^{\text{fro}})) > \mathcal{V}_{thr}
\label{eq:feature_condition}
\end{equation}
\begin{equation}
\sum_{c \in {FC}_i} \mathbb{I}(c \in \text{FOV}(v_{i,j}^{\text{fro}})) > {FC}_{thr}
\end{equation}
where \( {L} \) represents all currently known features, \( \mathbb{I}(\cdot) \) is an indicator function that equals 1 if the condition is met and 0 otherwise, and  \( \text{FOV}(\cdot) \) is the field of view function, which returns the set of points visible from a given viewpoint. The thresholds \( \mathcal{V}_{thr} \) and \( {FC}_{thr} \) denote the minimum number of features and frontier cells that must be visible for any viewpoint sampled. The number of features affects localization quality \cite{zhang2020fisher}, while the number of frontier cells ensures the exploration capability of sampled viewpoints \cite{zhou2021fuel}. An example of frontier-based viewpoints is shown in Fig.\ref{fig:viewpoint_graph}(b).

\textit{\textbf{Feature-based Viewpoints}}: Feature-based viewpoints are also considered into the graph, which are generated based on the following idea: all currently known features \( {L} \) can be clustered into several groups, resulting in feature clusters \( {LC}_1, {LC}_2, \dots, {LC}_N \), with each cluster containing enough features for localization. A feature-based viewpoint is expected to observe multiple feature clusters, such as \( {LC}_i \) and \( {LC}_j \). In this case, a transition from the viewpoint observing \( {LC}_i \) to that observing \( {LC}_j \) is ensured via the feature-based viewpoint, while maintaining localizability along the path.

Based on this idea, we sample viewpoints between any two feature clusters. Firstly, each cluster centroid \( o_i \) defines a candidate region \( \mathcal{R}_i \), constrained by a range \( [d_{\min}, d_{\max}] \), within which the UAV maintains sufficient visibility of the features in \( LC_i \). Then the feature-based viewpoint \( v_{i,j}^{\text{fea}}  = (p_{i,j}^{\text{fea}}, \psi_{i,j}^{\text{fea}}) \) is sampled from the overlapping region of \( \mathcal{R}_i \) and \( \mathcal{R}_j \), ensuring that it satisfies \eqref{eq:feature_condition} and:
\begin{equation}
\begin{aligned}
\left\{
\begin{array}{l}
p_{i,j}^{\text{fea}} \in \mathcal{R}_i \cap \mathcal{R}_j \\
o_i,o_j \in \text{FOV}(v_{i,j}^{\text{fea}})
\end{array}
\right.
\end{aligned}
\end{equation}
to ensure sufficient features for transition. An example of feature-based viewpoints is shown in Fig.\ref{fig:viewpoint_graph}(a).


\textit{\textbf{Viewpoint Transition Graph Construction}}: The graph is constructed from a set of frontier-based viewpoints \( \{v^{\text{fro}}\} \), feature-based viewpoints \( \{v^{\text{fea}}\} \), and the UAV's current state \( v_0 = (p_0, \psi_0) \). The edges between these viewpoints are defined as follows (see Fig.\ref{fig:viewpoint_graph}(c)):
\begin{enumerate}
    \item A unidirectional edge from \( v_0 \) to any \( v^{\text{fea}} \).
    \item A bidirectional edge between any two \( v^{\text{fea}} \).
    \item A unidirectional edge from any \( v^{\text{fea}} \) to any \( v^{\text{fro}}  \).
\end{enumerate}

Additionally, an edge is established between two viewpoints \( v_a \) and \( v_b \) only if their visible feature sets \( \Omega_a \) and \( \Omega_b \) satisfy:

\begin{equation}
\begin{aligned}
|\Omega_a \cap \Omega_b| > \mathcal{C}_{thr}
\label{eq:covisible_condition}
\end{aligned}
\end{equation}
where \( \mathcal{C}_{thr} \) represents the minimum number of co-visible features required for localization and the function \( |\cdot| \) denotes the cardinality of the set involved.

The transition cost between two viewpoints is defined as:

\begin{equation}
\begin{aligned}
C(v_a, v_b) = \max\left( \frac{\Vert p_a - p_b \Vert}{v_{\max}}, \frac{\Vert\psi_a - \psi_b\Vert}{\dot\psi_{max}} \right)
\label{eq:transition cost}
\end{aligned}
\end{equation}
where \( v_{\max} \) and \( \dot\psi_{max} \) denote the maximum translational and yaw velocities, respectively. The cost is the minimum time for both motions in parallel, set by the slower one.

\subsubsection{Viewpoint Evaluation and Selection}  
\label{subsec:viewpoint evaluation and selection}

Dijkstra searches are performed on the graph constructed in \ref{subsec:viewpoint transition graph}. Each search starts from \( v_0 \) and targets a frontier-based viewpoint in the set \( \{v^{\text{fro}}\} \) as the destination. A viewpoint is considered reachable while maintaining localizability if a valid path is found. The total cost of the found path is recorded, which represents the estimated minimum time \( t_{\text{sl}} \) required to reach \( v^{\text{fro}} \) from \( v_0 \), as derived from \eqref{eq:transition cost}. Similarly, using \eqref{eq:transition cost}, the time \( t_{\text{lg}} \) required to transition from \( v^{\text{fro}} \) to the goal (considering only the positional cost) can be estimated. The total transition time \( t_{\text{sg}} \) is then computed as:
\begin{equation}
t_{\text{sg}} = t_{\text{sl}} + t_{\text{lg}}
\end{equation}

We minimize \( t_{\text{sg}} \) to ensure time efficiency. In addition, viewpoint selection should ensure adequate exploration capability. The exploration gain \( \ell \) defined as the number of visible frontier cells, is used to quantify this capability. To further prioritize exploration in the direction of the goal, \( \ell \) is weighted by a navigation coefficient \( w \), as:
\begin{equation}
G_{\text{nav}} = w \cdot \ell = \frac{\mathbf{v}_{\text{goal}} \cdot \mathbf{v}_{\text{yaw}}}{\Vert\mathbf{v}_{\text{goal}}\Vert \Vert\mathbf{v}_{\text{yaw}}\Vert} \cdot \ell
\end{equation}
where \( \mathbf{v}_{\text{goal}} \) represents the vector from the current position to the final goal, and \( \mathbf{v}_{\text{yaw}} \) is the unit vector of the viewpoint’s yaw orientation. We refer to \( G_{\text{nav}} \) as the navigation gain.

The final reward function is defined as:
\begin{equation}
R = \omega_p \cdot (-t_{\text{sg}}) + \omega_n \cdot G_{\text{nav}}
\end{equation}
where \( \omega_p \) and \( \omega_n \) are the weights assigned to time efficiency and navigation gain, respectively. The viewpoint with the highest reward is selected as the next local target, and passed to the local planner with its associated frontier cells \( {FC}_e \).

\subsection{Position Trajectory Generation}
\label{subsec:position trajectory generation}
A position trajectory is generated to guide the UAV from the current state toward the given local target. The position trajectory $p(t)$ is parameterized as a degree $p=3$ uniform B-spline curve similar to \cite{zhou2019robust}, which is defined by control points $\mathbf{C}=\{\mathbf{C}_{0},\mathbf{C}_{1},\cdots,\mathbf{ C}_{M}\}$ with a time span $\Delta t$. 


Firstly, a kinodynamic A* algorithm is adopted to search for a collision-free path to the local target, and the search region is strictly limited to the known areas to avoid potential collisions. An initial B-spline curve is generated by fitting the search result, and the trajectory is subsequently refined through optimization that utilizes the convex hull property of B-splines. 
In order to generate a smooth, collision-free and dynamically feasible position trajectory while minimizing its duration to quickly reach the local target, the optimization problem is formulated as follows:
\begin{equation}
\begin{aligned}
\arg \min_{\mathbf{C},\Delta t} f_s + w_t t_{p} + \lambda_c f_c + \lambda_d (f_v + f_a)
\end{aligned}
\end{equation}
where \( f_s \) represents the smoothness cost, $t_{p}$ is the trajectory duration, \( f_c \) is the collision cost, and \( f_v \) and \( f_a \) are the velocity and acceleration penalty terms. \( \lambda_c \) and \( \lambda_d \) are the weights for the collision cost and dynamic feasibility terms.

\subsection{Yaw Trajectory Generation}
\label{subsec:yaw trajectory generation}

Traditional planning methods for quadrotor autonomous flight \cite{zhou2019robust}, \cite{zhou2020ego}, \cite{gao2023adaptive} employ head-forwarding yaw trajectories to rapidly explore unknown areas in the drone's velocity direction and avoid potential collisions along the trajectory. In contrast, perception-aware yaw trajectory generation methods \cite{zhang2020fisher}, \cite{murali2019perception}, \cite{chen2024apace} optimize the visibility of currently known features to reduce state estimation drift. This paper presents a yaw trajectory generation method that simultaneously considers exploration capability and localizability. The pipeline comprises three phases: initial path generation, localizable corridor construction, and trajectory optimization.


\subsubsection{Initial Path Generation}
\label{subsec:yaw initial path}

Inspired by \cite{zhou2021raptor}, the generation of the initial path is modeled as a graph search problem that seeks a sequence of yaw angles $\Omega_{\psi}:=\{(\psi_i)_{i=0:N}\}$. Firstly, uniform time-interval sampling with $t_{\psi}$ is applied to the position trajectory and its second order derivatives, yielding discrete positions $\Omega_{P}: = \{(p_i)_{i=0:N}\}$ with corresponding accelerations $\Omega_{\mathbf{a}}:=\{(\mathbf{a}_i)_{i=0:N}\}$. In order to avoid too long sample intervals, $t_{\psi}$ is calculated as follows:

\vspace{-0.2cm}

\begin{equation}
\begin{aligned}
t_{\psi}=t_{p}\cdot\left(\lfloor \frac{t_{p}}{t_{f}} \rfloor +1\right)^{-1}
\label{eq:t_yaw}
\end{aligned}
\end{equation}
where $ \lfloor\cdot\rfloor $ represents the floor function, and $t_{f}$ represents the maximum tolerance duration for VIO to localize accurately by integrating inertial measurements under visually degraded conditions. Uniform yaw angle sampling is carried out at each intermediate position $p_i \in \Omega_{P},1\leq i \leq N-1$. For each sampled $\psi_{i,j}$, a corresponding node $\mathcal{N}(p_i,\mathbf{a}_i,\psi_{i,j})$ is instantiated and its 6-DOF pose $T_{i,j}$ is derived based on the differential flatness property to achieve more accurate observation. We denote $\mathcal{N}(p_i,\mathbf{a}_i,\psi_{i,j})$ as $\mathcal{N}_{i,j}$ for simplicity without ambiguity. $\mathbb{L}_{i,j}$ and $\mathbb{F}_{i,j}$ are respectively denoted as the sets of indices for features and cells of $FC_{e}$ (provided by \ref{subsec:viewpoint generation}) that are visible to $\mathcal{N}_{i,j}$. The algorithm rejects adding $\mathcal{N}_{i,j}$ to the graph if $\left| \mathbb{L}_{i,j} \right| \leq \mathcal{V}_{thr}$, where $\mathcal{V}_{thr}$ is defined as \eqref{eq:feature_condition}. We name it the \textit{visibility constraint} of nodes.

In particular, we directly generate nodes $\mathcal{N}_0$ and $\mathcal{N}_N$ from the current estimated state and the local target state (provided by \ref{subsec:viewpoint generation}) instead of sampling.

Edges are established between adjacent-position nodes $\mathcal{N}_{i,j}$ and $\mathcal{N}_{i+1,k}$. Besides the smoothness constraint $\Vert\psi_{i,j}-\psi_{i+1,k}\Vert \textless \dot\psi_{max} \cdot t_{\psi}$, the algorithm checks the number of co-visible features between them, and rejects connecting them if $\left| \mathbb{L}_{i,j} \cap \mathbb{L}_{i+1,k} \right| \leq \mathcal{C}_{thr}$, where $\mathcal{C}_{thr}$ is defined as \eqref{eq:covisible_condition}. We name it the \textit{co-visibility constraint} of edges. 

\begin{figure}[tp]
    \centering
    \includegraphics[width=\columnwidth]{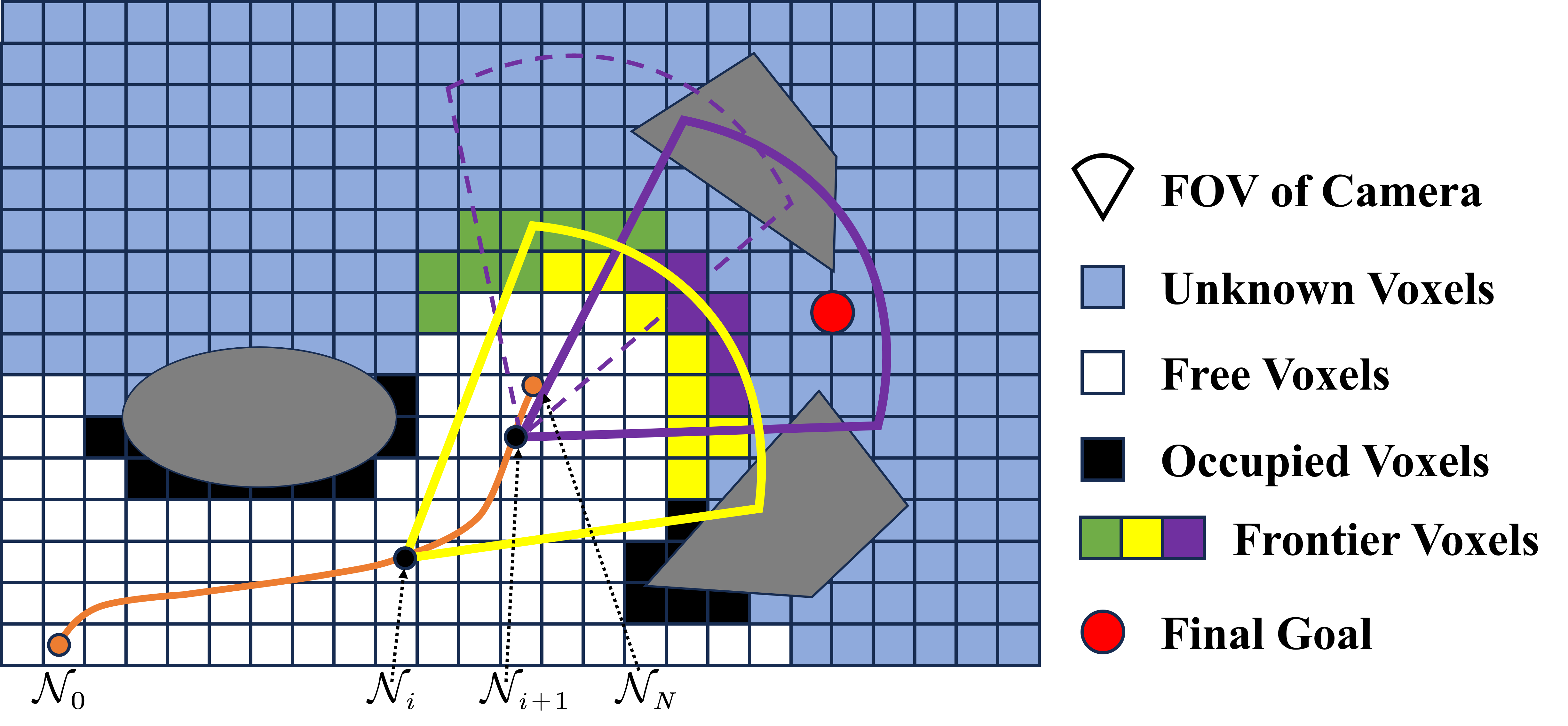}
    \caption{Illustration of $c_{e}(\mathcal{N}_i,\mathcal{N}_{i+1})$. Voxels in yellow, purple, green respectively represent $\mathbb{F}_i,\mathbb{F}_{i+1}\setminus\mathbb{F}_i,FC_e\setminus(\mathbb{F}_i\cup\mathbb{F}_{i+1})$. Only the purple voxels are included in the calculation of $c_{e}(\mathcal{N}_i,\mathcal{N}_{i+1})$. Additionally, compared with another node (represented by purple dashed line), $\mathcal{N}_{i+1}$ is more conducive to expanding the known region to the final goal to complete the navigation as soon as possible, which is rewarded by the goal-directed term in $c_{e}(\mathcal{N}_i,\mathcal{N}_{i+1})$.}
    \label{fig:see_final_goal}
    \vspace{-0.4cm}
\end{figure}

After the graph construction, we aim to search a minimum cost path $\mathcal{S}^{*}:=\{(\mathcal{N}_i)_{i=0:N}\}$ and extract $\Omega_{\psi}$. The overall cost function $c_s$ is defined as: 
\begin{equation}
\begin{aligned}
&c_s(\mathcal{S}) = \sum_{i = 0}^{N-1} \{(\psi_{i+1}-\psi_{i})^2 \cdot \left[ 1+\exp(-\mu_{e}c_{e}(\mathcal{N}_{i},\mathcal{N}_{i+1})) \right] \}\\
&c_{e}(\mathcal{N}_{i},\mathcal{N}_{i+1}) =  \mu_{f}\left| \mathbb{F}_{i+1} \setminus \mathbb{F}_{i} \right| + \mu_{g}\mathbb{I}(\mathbf{p}_{goal} \in \text{FOV}(\mathcal{N}_{i+1}))\\
\end{aligned}
\end{equation}
where $\mu_{e},\mu_{f},\mu_{g} \in\mathbb{R}^+$ are penalty scale factors, and $\mathbf{p}_{goal}$ represents the final goal. The weight of the edge is set according to its smoothness and navigation gain as shown in Fig.\ref{fig:see_final_goal}. The first term of $c_{e}(\mathcal{N}_{i},\mathcal{N}_{i+1})$ is defined according to the number of newly visible cells of $FC_{e}$ when transitioning from $\mathcal{N}_{i}$ to $\mathcal{N}_{i+1}$, and its second term is defined to enhance the goal-directed perception. Finally, a Dijkstra algorithm is employed to effectively solve the graph search problem.

\subsubsection{Trajectory Optimization}
\label{subsec:yaw trajectory optimization}
Nonlinear optimization is employed to generate a smooth, dynamically feasible and localizable trajectory with $\Omega_\psi$ as initial values. In our work, the localizability is ensured by constructing a series of angular boundary pairs as constraints which will be introduced in \ref{subsec:yaw localizable corridor generation}. The angular boundary pairs are denoted as $(\mathcal{L},\mathcal{U}):=\{ ( \mathfrak{l}_{i},\mathfrak{u}_{i})_{i=1:N-1}\}$, and for any optimization variable $\psi_i^{'}$, it's constrained with $\psi_i^{'} \in (\mathfrak{l}_i,\mathfrak{u}_i)$. In order to efficiently solve this constrained optimization problem, a similar method to \cite{wang2022geometrically} is employed to transform it into an equivalent unconstrained optimization problem defined on $\mathbb{R}$. The yaw trajectory $\psi \left( t \right)$ is  parameterized as a $\mathfrak{T}_{\text{MINCO}}^3$ trajectory \cite{wang2022geometrically}, determined by intermediate points $\mathbf{Q}:=\{(q_i)_{i=1:N-1}\}$ and fixed time vector $\mathbf{T}:=\{(t_i=t_{\psi})_{i=1:N-1}\}$. The following formula is adopted to map $\psi_i$ to the corresponding $q_i\in\mathbb{R}$:

\begin{equation}
\begin{aligned}
r_i=\frac{\psi_i-\mathfrak{l}_i}{\mathfrak{u}_i-\mathfrak{l}_i},\quad q_i=-\log(\frac{1-r_i}{r_i})
\end{aligned}
\end{equation}

Thanks to the explicit diffeomorphism, the cost functions $J_{\psi}$ defined below, as well as its gradient, can be directly transformed into $J_{q}$ and $J_{q}^{'}$. We refer the readers to \cite{wang2022geometrically} for more details. The overall cost function is defined as:

\begin{equation}
\begin{aligned}
J_{\psi}=\sum_{i = 1}^{N-1}\left[J_{s}(\psi_i)+\lambda_{d} J_{d}(\psi_i)
-\lambda_{e} J_{e}(\psi_i)\right]
\label{equation:yaw opti}
\end{aligned}
\end{equation}
where $\lambda_{d},\lambda_{e} \in\mathbb{R}^+$ are adjustable weights. $J_{s}(\psi_i)$ and $J_{d}(\psi_i)$ are designed to be similar to \cite{zhou2020ego}, which account for smoothness and dynamic feasibility, respectively. $J_{e}(\psi_i)$ represents the exploration capability of $FC_{e}$ and we calculate it as the sum of the visibilities of multiple cells $v(\mathbf{c}_j,\mathbf{s}(\psi_i))$. Here, $\mathbf{c}_j$ represents the $j$-th cell of $FC_e$ and the 6-DOF pose $\mathbf{s}(\psi_i)$ is derived similar to \ref{subsec:yaw initial path}. The differentiable visibility metric proposed in \cite{chen2024apace} is leveraged to quantify $v(\mathbf{c}_j,\mathbf{s}(\psi_i))$. The higher the value of $v(\mathbf{c}_j,\mathbf{s}(\psi_i))$, the closer $\mathbf{c}_j$ is to the center of the FOV of $\mathbf{s}(\psi_i)$. We refer the readers to \cite{chen2024apace} for the specific definition of $v(\mathbf{c}_j,\mathbf{s}(\psi_i))$.

Prior to optimizing $\psi_i$, the data stored in $\mathcal{N}_{i}$ in \ref{subsec:yaw initial path} is leveraged to filter out those cells in $FC_e$ that 
are impossible to be visible or that make no contribution to enhancing the explore capability. Specifically, for $\mathbf{c}_j \in FC_{e}$, it does not participate in the calculation of $J_{e}(\psi_{i})$ if and only if:
\begin{itemize}{\leftmargin=1em}
    \setlength{\topmargin}{0pt}
    \setlength{\itemsep}{0pt}
    \setlength{\parsep}{0pt}
    \setlength{\parskip}{0pt}
\item Raycast from $\mathbf{c}_j$ to $\mathcal{N}_{i}$ is blocked by obstacles. 
\item $\mathbf{c}_j$ has been observed by previous nodes $\{ ( \mathcal{N}_j)_{j<i}\}$
\end{itemize}

Denote $\mathcal{X}_i \in FC_e$ as the set of cells after filtering, and $J_{e}(\psi_i)$ is defined as $J_{e}(\psi_{i})=\sum_{\mathbf{c} \in \mathcal{X}_i}v(\mathbf{c},\mathbf{s}(\psi_{i}))$.

\subsubsection{Localizable Corridor Construction}
\label{subsec:yaw localizable corridor generation}
The construction of the localizable corridor is modeled as a problem that seeks a sequence of angular boundary pairs $(\mathcal{L},\mathcal{U})$ such that the localizability is ensured for all feasible solutions of the constrained optimization problem constructed from $(\mathcal{L},\mathcal{U})$ in \ref{subsec:yaw trajectory optimization}. We name these $(\mathcal{L},\mathcal{U})$ the \textit{localizable corridor}.


One way to solve the above problem is as follows. Given the initial path $\mathcal{S}:=\{(\mathcal{N}_i)_{i=0:N}\}$ and corresponding $\Omega_{\psi}:=\{(\psi_i)_{i=0:N}\}$, the visible feature sets of each node $\mathbb{L}:=\{(\mathbb{L}_i)_{i=0:N}\}$ ensure the visibility and co-visibility constraints defined in \ref{subsec:yaw initial path} of the entire path. Then, for each intermediate node $\mathcal{N}_i \in \mathcal{S},1\leq i \leq N-1$, we can extract a critical feature subset $\mathbb{L}_i^{\text{cri}} \subseteq \mathbb{L}_i$ and search for a pair of angular boundaries $(\mathfrak{l}_i,\mathfrak{u}_i)$ such that:
\begin{equation}
\begin{aligned}
\left\{
\begin{array}{l}
\psi_i\in(\mathfrak{l}_i,\mathfrak{u}_i)\\
\forall \psi^{'} \in (\mathfrak{l}_i,\mathfrak{u}_i),\mathbf{f}\in\mathbb{L}_i^{\text{cri}},\mathbf{f} \in \text{FOV}(\mathcal{N}(p_i,\mathbf{a}_i,\psi^{'}))
\end{array}
\right.
\end{aligned}
\end{equation}

If the selected $\mathbb{V}:=\{(\mathbb{L}_i^{\text{cri}})_{i=1:N-1}\}$ still satisfies the visibility and co-visibility constraints, corresponding $(\mathcal{L},\mathcal{U}):=\{ ( \mathfrak{l}_{i},\mathfrak{u}_{i})_{i=1:N-1}\}$ is a solution to the above problem. In this way, the original problem is transformed into the problem of selecting a sequence of critical feature subsets.

\begin{figure}[tp]
    \centering
    \includegraphics[width=\columnwidth]{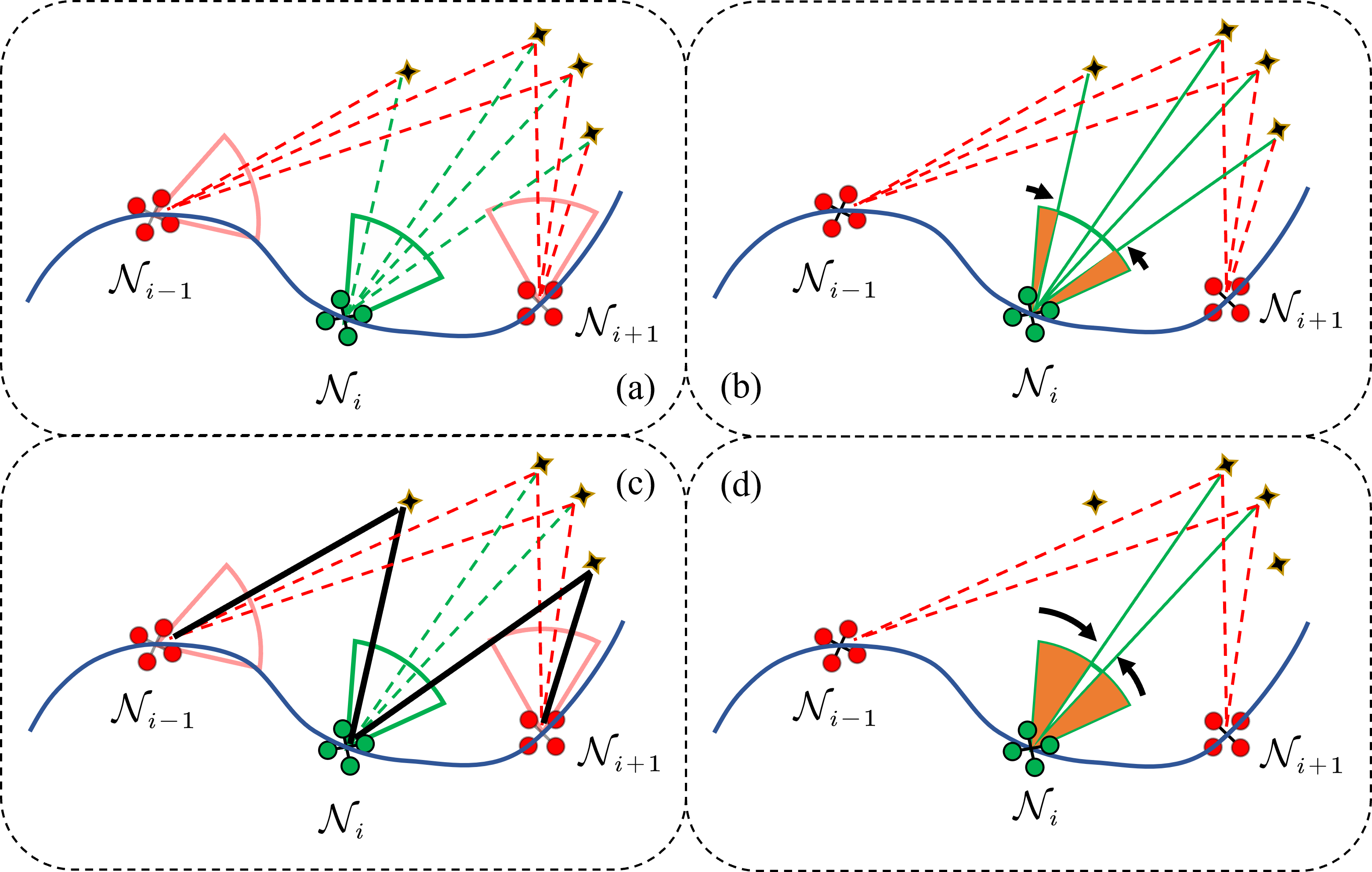}
    \caption{Select $\mathbb{L}_i^{\text{cri}}$ during the corridor construction. (a) Co-visible features represented with stars between $\mathcal{N}_i$ and its adjacent nodes. (b) Select $\mathbb{L}_i^{\text{cri}}=\mathbb{L}_i$ and lead to a narrow corridor. (c) Co-visibility evaluation. The features with black solid line are evaluated with poor co-visibility. (d) Select features with high co-visibility from $\mathbb{L}_{i}$ to form $\mathbb{L}_{i}^{\text{cri}}$ and generate a significantly wider corridor than (b).}
    \label{fig:fuck_yaw_corridor}
     \vspace{-0.4cm}
\end{figure}

An overly conservative selection strategy, such as the one that maximizes the retention of localizability, will lead to a narrow corridor and severely restrict the feasible region of optimization as shown in Fig.\ref{fig:fuck_yaw_corridor}(b). In order to widen the coverage of the generated corridor as much as possible, our method performs feature co-visibility analysis on adjacent nodes leveraging the visual model mentioned in \ref{subsec:yaw trajectory optimization} as shown in Fig.\ref{fig:fuck_yaw_corridor}(c). Denote $\mathbb{L}_c=\mathbb{L}_{i} \cap \mathbb{L}_{i-1}$ as the co-visible feature set between $\mathcal{N}_{i}$ and $\mathcal{N}_{i-1}$. For any $\mathbf{f} \in \mathbb{L}_c$, its co-visibility measure with respect to $\mathcal{N}_{i}$ and $\mathcal{N}_{i-1}$ is defined as $\mu(\mathbf{f},\mathcal{N}_{i-1},\mathcal{N}_{i})=v(\mathbf{f},T_{i-1})\cdot v(\mathbf{f},T_i)$. After that, the top $\mathcal{C}_{thr}$ features with the highest co-visibility in $\mathbb{L}_c$ are extracted to obtain the first set $\mathcal{K}_{1}$. The same procedure is applied to $\mathcal{N}_{i}$ and $\mathcal{N}_{i+1}$ to obtain $\mathcal{K}_{2}$. Finally, $\mathcal{K}_{1}$ and $\mathcal{K}_{2}$ are merged to obtain $\mathbb{L}_{i}^{\text{cri}}$. In this way, we not only consider the co-visibility constraints between $\mathcal{N}_i$ and its adjacent nodes simultaneously to ensure the localizability, but also retain those co-visible features that are located at both the centers of the FOV of the adjacent nodes as much as possible, so as to generate a wider corridor as shown in Fig.\ref{fig:fuck_yaw_corridor}(d).

Following the selection of $\mathbb{L}_i^{\text{cri}}$, an incremental bidirectional search method is employed to generate $(\mathfrak{l}_i,\mathfrak{u}_i)$ from $\mathbb{L}_i^{\text{cri}}$. The algorithm starts with $\psi_i$ and increments in both clockwise and counterclockwise directions. In a single direction, denote the angle of the k-th iteration as $\psi_{k}^{it}$ ($\psi_{0}^{it}=\psi_i$), and instantiate $\mathcal{N}_{k}^{it}$ similar to \ref{subsec:yaw initial path}. When the visibility constraint of $\mathcal{N}_{k}^{it}$ is no longer satisfied or $\exists \mathbf{f} \in \mathbb{L}_i^{\text{cri}}, \mathbf{f} \notin \text{FOV}(\mathcal{N}_k^{it})$, the algorithm stops iterating, and $\psi_{k-1}^{it}$ is recorded as the boundary found in this direction.

%% file: sections/results.tex
\section{Experiments}
\label{sec:experiments}



\subsection{Simulation Benchmarks}
\label{subsec:simulation benchmarks}

We perform a comparison with Zhang's method \cite{zhang2018perception} and APACE \cite{chen2024apace}. Since APACE is not designed for navigation in unknown environments, we introduce two adapted versions for our simulation scenarios:

\begin{itemize}
\item 
APACE-R: Integrating APACE with a perception-agnostic local replanning strategy similar to \cite{zhou2019robust}.

\item 
APACE-E: Integrating APACE with our adaptive local target selection method described in \ref{subsec:viewpoint generation} to enhance its exploration capability. 

\end{itemize}

\begin{figure*}[t]
    \centering
    \includegraphics[width=17.5cm]{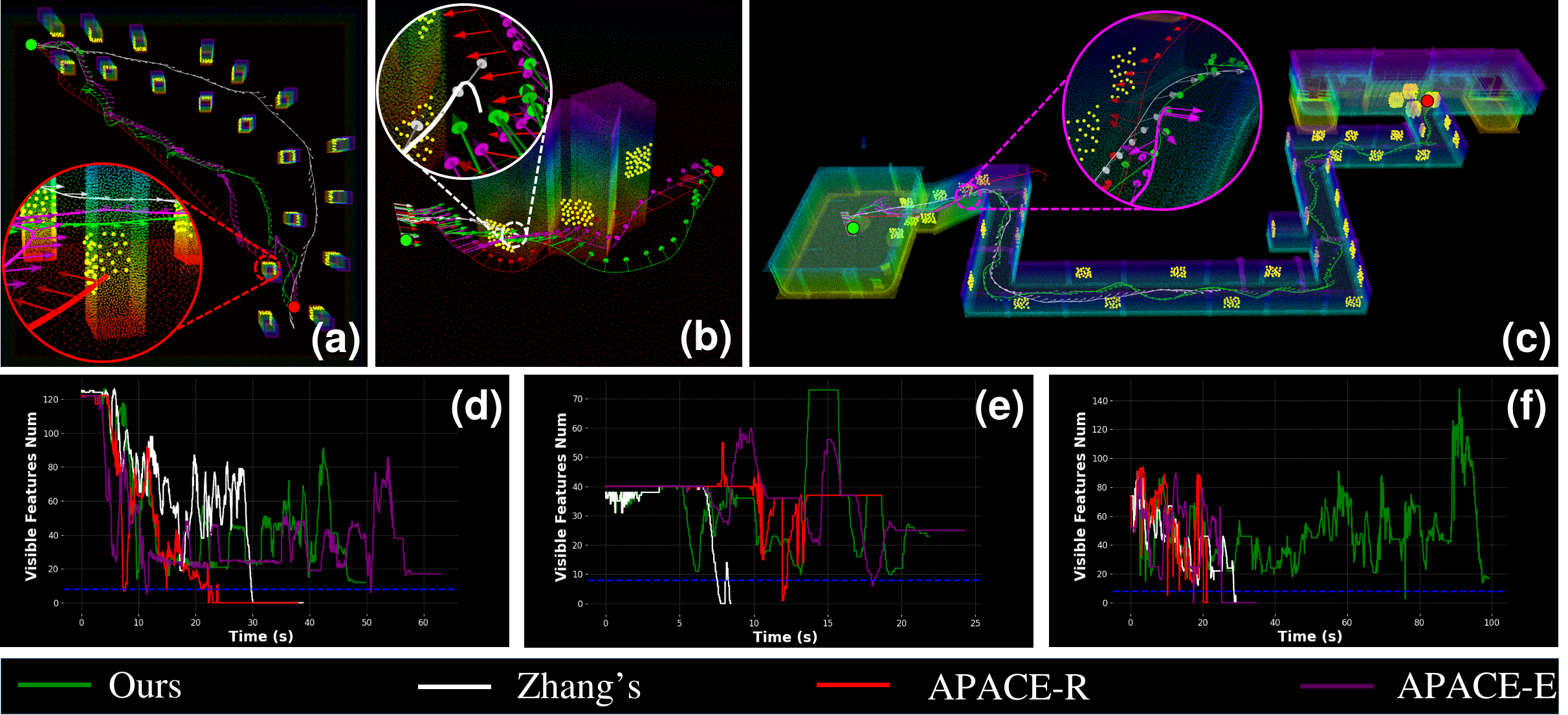}  

     \caption{Comparison of the four methods across different scenarios (a), (b), and (c), where green, red, and yellow dots respectively indicate the start positions, the goal positions, and the features. Each of these subfigures includes zoomed-in sections highlighting critical details: (a) APACE-R colliding with an obstacle, (b) Zhang's method failing to observe sufficient features, and (c) APACE-E generating a dynamically infeasible trajectory. (d), (e), and (f) show the visible feature count over time for each algorithm in the corresponding scenarios. The UAV is considered to have lost localization if visible features remain below a threshold (blue dashed line) for longer than $t_f$.}
    \label{fig:benchmark}
    \vspace{-0.5cm}
\end{figure*}

We conduct three scenarios as shown in Fig.\ref{fig:benchmark}. The scenarios in Fig.\ref{fig:benchmark}(a)(b) are adapted from the simulation environments in \cite{zhang2018perception} and \cite{chen2024apace}. Additionally, a long corridor scenario, shown in Fig.\ref{fig:benchmark}(c), is designed to compare the navigation efficiency and localization capability of different algorithms in more complex cases, including sharp turns, dead-ends, and staircases with limited features. Each experiment involves navigating from the start position to the final goal without prior environmental knowledge, repeated over 100 runs. The key parameters are set as follows: \( v_{\max} = 1.5 \) m/s, \(\mathbf{a}_{\max} = 1.5 \) m/s\(^2\), and \( t_f = 0.35 \)s (defined in \ref{subsec:yaw initial path}).





The following indicators are employed to compare our method with baselines: success rate (the definition of mission failure has been described in \ref{sec:problem_formulation}), the average navigation time (only conducted when the mission is successful), and the average algorithm runtime (maximum computation time of single trajectory generation). 


The quantitative results of these indicators are presented in Table \ref{tab:benchmark}. Zhang’s method and APACE-R achieve a certain success rate only in scenarios similar to those in their respective studies. Zhang’s method follows a fixed head-forward yaw trajectory, failing to adjust the yaw for localization in Scenes (b) and (c). APACE-R lacks exploration capability, causing the FOV to be restricted by visual features and overlook obstacles, which results in collisions in Scenes (a) and (c). 

APACE-E, incorporating a local target selection strategy with exploration capability, achieves a certain success rate in Scenes (a) and (b). However, it repeatedly evaluates the visibility of numerous features during the trajectory generation, which leads to long replanning time and poor navigation efficiency. Additionally, its yaw trajectories overemphasize localizability and fail to guarantee the dynamic feasibility, as observed in Scene (c), causing the controller to fail in tracking.  


\begin{table}[hbtp]
\small
\caption{Performance comparison of our method and baselines}
\label{tab:benchmark}
\begin{tabular}{ccccc}
\toprule
\toprule
\textbf{Scenario} & \textbf{Method} & \textbf{\makecell[c]{Success\\Rate}}  & \textbf{\makecell[c]{Navigation\\Time(s)}} & \textbf{\makecell[c]{Algorithm\\Runtime(ms)}} \\
\midrule
Scene(a) & 
\makecell[c]{Zhang's\\
APACE-R\\APACE-E
\\Ours}  &
\makecell[c]{23/100\\0/100\\
42/100\\\textbf{100/100}} & 
\makecell[c]{\textbf{38.26}\\-\\
62.87\\42.41} &
\makecell[c]{102.60\\408.33\\
391.04\\\textbf{45.21}} \\
\midrule
Scene(b) & 
\makecell[c]{Zhang's\\
APACE-R\\APACE-E\\Ours} & 
\makecell[c]{0/100\\49/100\\
\textbf{100/100}\\\textbf{100/100}} & 
\makecell[c]{-\\18.69\\
30.80\\\textbf{15.76}} &
\makecell[c]{54.17\\
301.78\\201.44\\\textbf{23.12}} \\
\midrule
Scene(c) & 
\makecell[c]{Zhang's\\
APACE-R\\APACE-E\\Ours} & 
\makecell[c]{0/100\\0/100\\
0/100\\\textbf{92/100}} & 
\makecell[c]{-\\
-\\-\\\textbf{110.95}} &
\makecell[c]{81.41\\772.66\\
472.90\\\textbf{62.33}}\\
\bottomrule
\bottomrule

\end{tabular}
\vspace{-0.5cm}
\end{table}

Compared to the baselines, our method balances localization and navigation efficiency in local target selection while ensuring that each local trajectory is collision-free, dynamically feasible, and localizable. As a result, our method achieves the highest success rate and the lowest algorithm runtime across all scenarios with acceptable navigation time.  


\subsection{Ablation Study on Viewpoint Transition Graph}
\label{subsec:ablation_study_on_viewpoint_transition_graph}
An ablation experiment is designed to evaluate the contribution of the viewpoint transition graph in \ref{subsec:viewpoint transition graph} to overall navigation performance. Specifically, we remove the graph search module from the original system and assume that each frontier-based viewpoint is directly accessible.

\begin{figure}[hb]
    \centering
    \includegraphics[width=\columnwidth]{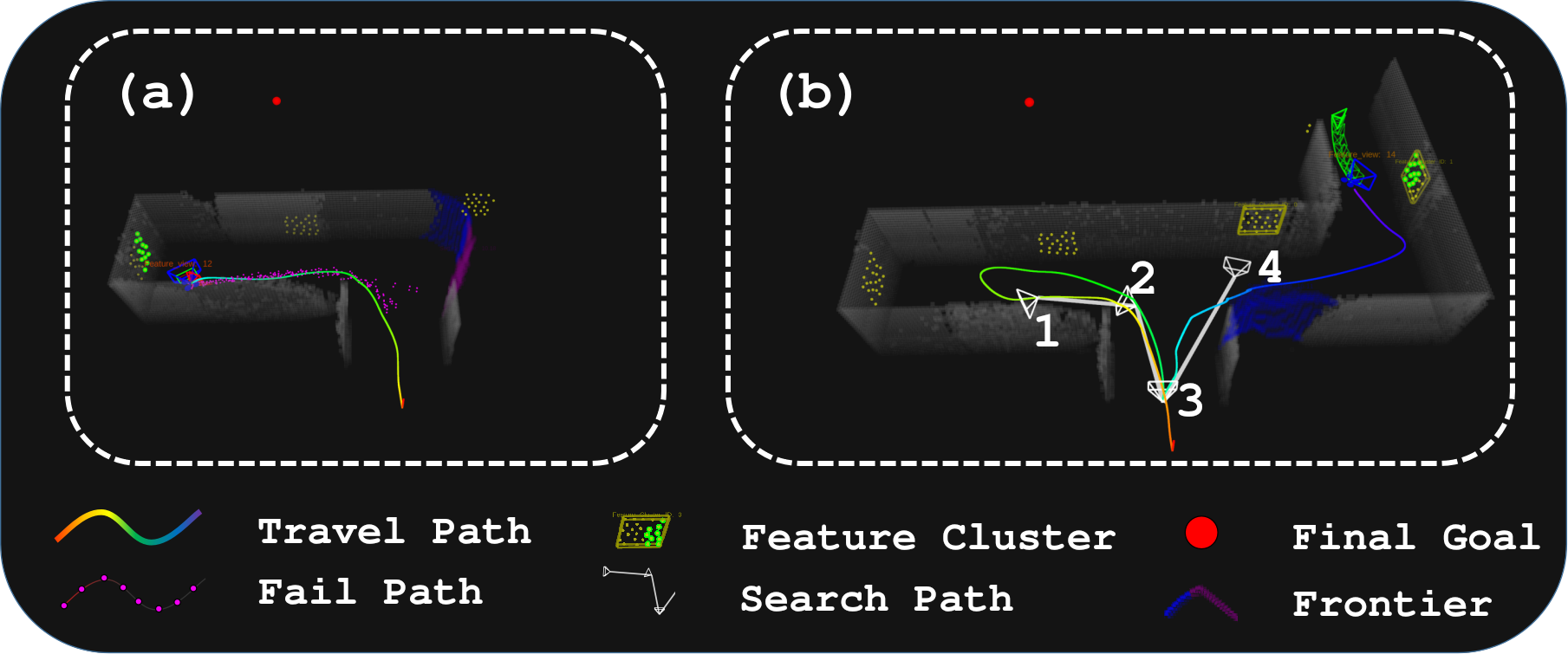}
    \caption{Ablation study on the viewpoint transition graph. (a) Without the graph, the UAV gets trapped in a dead-end. (b) With the graph, the planning method generates localizable trajectories, allowing the UAV to navigate out of the dead-end successfully.}
    \label{fig:ablation_study}
    \vspace{-0.55cm}
\end{figure}

A dead-end scenario is introduced for the experiment. Without the graph search module, the generated position trajectories lack sufficient co-visual features along the path, making yaw trajectory generation infeasible. Consequently, the co-visibility constraint in \ref{subsec:yaw initial path} cannot be satisfied along the entire path, as shown in Fig.\ref{fig:ablation_study}(a). In contrast, our system constructs a viewpoint transition graph based on observed features, guiding the generation of a smooth and localizable trajectory. This allows the UAV to successfully navigate out of the dead-end as shown in Fig.\ref{fig:ablation_study}(b).

\subsection{Real-world Experiments}
\label{subsec:real-world experiments}

A series of real-world experiments are conducted to further verify the feasibility of the proposed method. A custom-built quadrotor platform is employed for the experiments, which is equipped with an Intel NUC featuring an Intel Core i7-1260P CPU and 32GB memory, along with a forward-looking camera RealSense D455. The state estimation module is based on ORB-SLAM3 \cite{campos2021orb}, operating in RGB-D mode. In addition, an OptiTrack motion capture system is deployed to provide accurate ground-truth odometry data and quantify state estimation drift errors. Notice that onboard lighting is used only for photography and does not illuminate obstacles. 

\begin{figure}[tbp]
    \centering
    \includegraphics[width=\columnwidth]{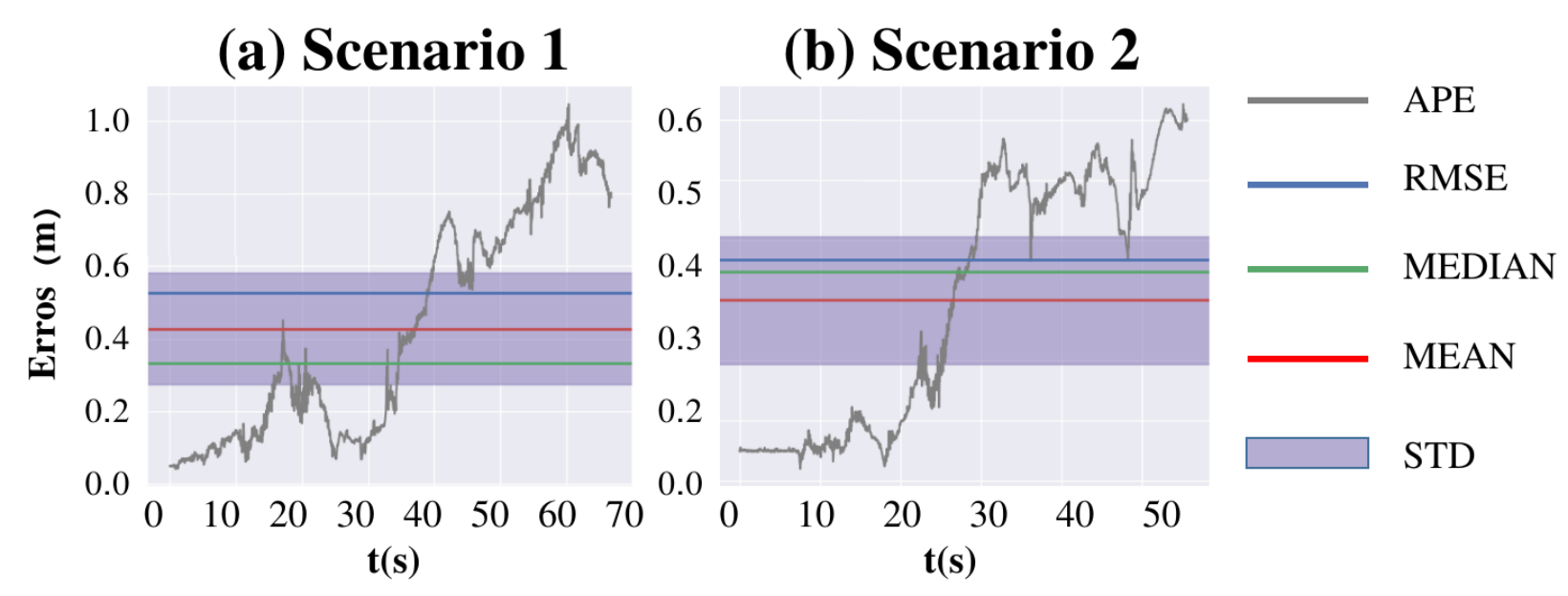} 
    \caption{Error analysis for two different scenarios. The curves show the error between odometry data and ground truth. APE stands for the Absolute Positional Error, while RMSE, MEDIAN, MEAN, and STD summarize the overall error distribution.}
    \label{fig:errors_curves}
    \vspace{-0.5cm}
\end{figure}

We design two dark scenarios where the UAV is required to navigate to a target located 12 meters away. In the first scenario shown in Fig.\ref{fig:real}, the UAV encounters a previously unknown dead-end. At this point, labeled as 2, its FOV contains only a sparse set of features, making it difficult to generate a path that maintains localizability while escaping. Our algorithm successfully navigates out by searching for reachable viewpoints that enable transitions to a local target with sufficient localization. It efficiently completes the navigation task with 27.66m path length, 0.79m goal error and 0.53m RMSE for state estimation, as shown in Fig.\ref{fig:errors_curves}(a).

\begin{figure}[hbt]
    \centering
    \includegraphics[width=\columnwidth]{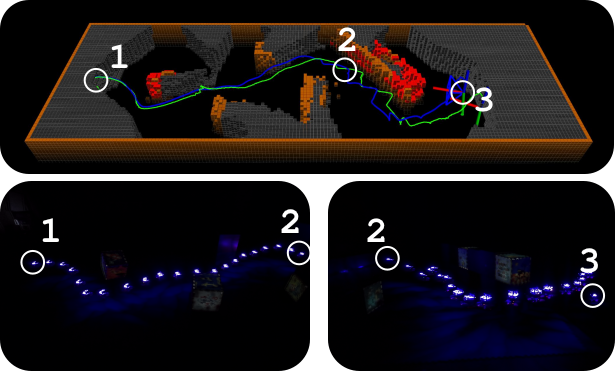} 
    \caption{The UAV navigates through numerous obstacles, only some of which emit light, making it challenging to allocate perception resources between feature observation and unknown area detection.}
    \label{fig:real_simple_scene}
    \vspace{-0.3cm}
\end{figure}

In the second scenario shown in Fig.\ref{fig:real_simple_scene}, numerous obstacles are placed in a dark environment, with only some emitting light. Since the UAV cannot extract enough features for localization in dark areas, it needs to focus on observing illuminated boxes to maintain localization. Meanwhile, perception resources must also be allocated to detecting obstacles in the unknown environment to ensure safe navigation and avoid collisions. Through replanning, the UAV completes a total path length of 19.21m while avoiding collisions, with a goal error of 0.60m and a state estimation RMSE of 0.37m, as shown in Fig.\ref{fig:errors_curves}(b).

The experimental results demonstrate that our method enables gradual exploration in unknown and feature-limited environments while maintaining localization and efficiently navigating. More details are in the submitted video.

%% file: sections/conclusion.tex
\section{Conclusion}
\label{sec:conclusion}


This paper presents a perception-aware planning method for quadrotor flight in unknown and feature-limited environments. Our approach establishes a systematic mechanism to properly allocate perception resources among environmental information in a computationally efficient way. Firstly, a viewpoint transition graph is proposed to adaptively select local targets, balancing localization and navigation efficiency. Besides, at the trajectory generation level, a localizable corridor-based method is proposed to ensure the localization robustness efficiently without numerous evaluation of visibility during the optimization. Our approach enables efficient perception-aware navigation without any prior environmental knowledge. The feasibility of our method has been validated through both simulations and real-world experiments.